\documentclass[10pt,twocolumn,letterpaper]{article}

\usepackage{cvpr}
\usepackage{times}
\usepackage{epsfig}
\usepackage{graphicx}
\usepackage{amsmath}
\usepackage{amssymb}

\usepackage[pagebackref=true,breaklinks=true,letterpaper=true,colorlinks,bookmarks=false]{hyperref}

\cvprfinalcopy 

\def\cvprPaperID{7850} 

\ifcvprfinal\fi

\begin{document}

\title{VSGNet: Spatial Attention Network for Detecting Human Object Interactions Using Graph Convolutions\vspace{-0.3cm}}

\author{Oytun Ulutan\thanks{Authors Contributed Equally}\hspace{0.1cm} \thanks{Ulutan is currently with the Vision team of Zoox, Inc.}
\qquad\qquad A S M Iftekhar\textsuperscript{*}\qquad\qquad B. S. Manjunath\\ Department of Electrical and Computer Engineering\\University of California, Santa Barbara, CA}

\maketitle
\thispagestyle{empty}

\begin{abstract}
  Comprehensive visual understanding requires detection frameworks that can effectively learn and utilize object interactions while analyzing objects individually. This is the main objective in Human-Object Interaction (HOI) detection task. In particular, relative spatial reasoning and structural connections between objects are essential cues for analyzing interactions, which is addressed by the proposed Visual-Spatial-Graph Network (VSGNet) architecture. VSGNet extracts visual features from the human-object pairs, refines the features with spatial configurations of the pair, and utilizes the structural connections between the pair via graph convolutions. The performance of VSGNet is thoroughly evaluated using the Verbs in COCO (V-COCO) and HICO-DET datasets. Experimental results indicate that VSGNet outperforms state-of-the-art solutions by 8\% or 4 mAP in V-COCO and 16\% or 3 mAP in HICO-DET. Code is available online.\footnote{\href{https://github.com/ASMIftekhar/VSGNet}{https://github.com/ASMIftekhar/VSGNet} } \vspace{-0.5cm} 

\end{abstract}

\section{Introduction}

The task of detecting human object interaction (HOI) in images refers to detecting the interactions between a human and object pair and localizing them. HOI detection can be considered a part of the task of visual scene understanding ~\cite{eslami2016attend,zhou2019semantic,xiao2018unified}, visual question answering\cite{fan2018stacked,yu2017multi,nam2017dual,teney2018tips}, and activity recognition in videos ~\cite{liu2019caesar,wu2019learning,ulutan2018actor}. Although there has been significant improvements in recent years for detecting and recognizing objects ~\cite{girshick2014rich,ren2015faster, he2016deep}, HOI detection still poses various challenges. For example, interactions usually happen in a subtle way, same types of relations vary significantly across different settings, multiple humans can interact with the same object or vice-versa, and different relations might have visually subtle differences ~\cite{gupta2015visual,chao2018learning}.


\begin{figure}[t]
\begin{center}
\includegraphics[width=1.0\linewidth]{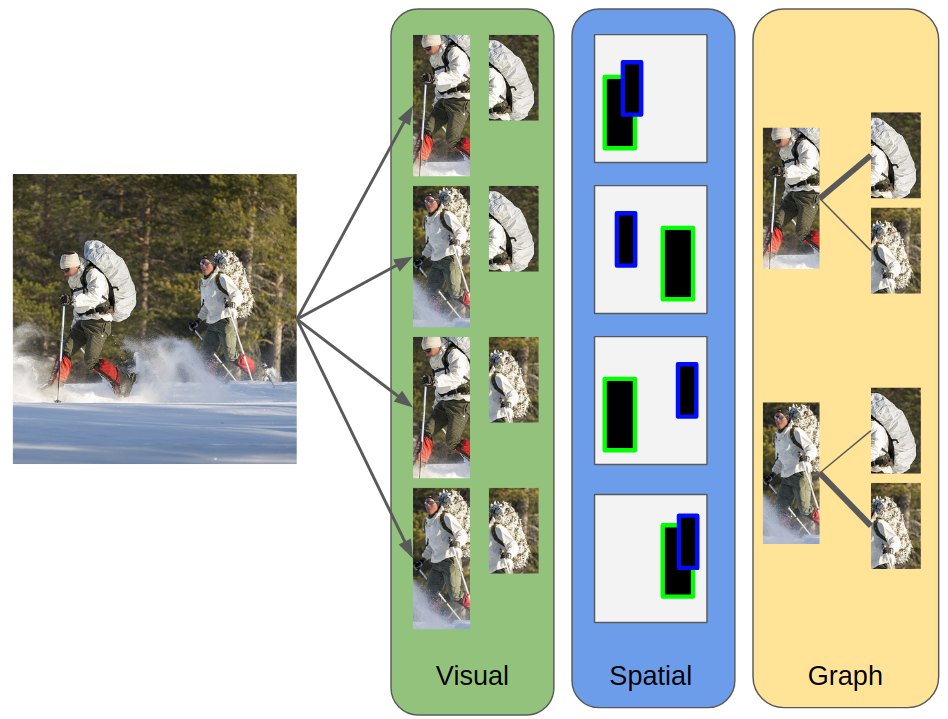}
\end{center}
   \caption{ Visual, Spatial and Graph branches of our proposed VSGNet model. Visual branch analyzes humans/objects/context individually, Spatial branch uses spatial configurations of the pairs to refine visual features and the Graph branch utilizes the structural connections by Graph convolutions which uses interaction proposal scores as edge intensities between human-object nodes. \vspace{-0.5cm}
   }
\label{fig:three_branches}
\end{figure} 

Most of the existing methods in HOI detection task~\cite{li2019transferable, gao2018ican,gkioxari2018detecting} follow a similar structure. Using an object detection framework, human and object features are extracted. These features are paired exhaustively along with some other features (e.g. pose, relative geometric locations) ~\cite{li2019transferable,gao2018ican} and then fed into a multi-branch deep neural network to detect the relationship between humans and objects. 
Even though this approach achieves good results for detecting HOIs, it does not explicitly utilize the interaction information or the spatial relations between pairs.
HOIs such as person on a skateboard or a person holding a bat have well defined spatial relations and structural interactions which should be leveraged in this detection task. 


For utilizing spatial configurations, VSGNet uses a spatial attention branch that explicitly uses the spatial relations of the pairs to refine the visual features. 
Instead of modeling humans and objects individually, our attention module uses the spatial configurations of the pairs to extract attention weights which refine the visual features. Although a few past works ~\cite{chao2018learning,gao2018ican} have used these types of spatial configurations as features for classification directly, these models do not combine the visual information with spatial information. These features are more useful for refining the visual features and providing an attention mechanism for modeling the interactions of the human-object pairs explicitly.


For modeling the interactions, an image can be defined as a graph. Nodes in this graph are the humans and objects, in which case the edges define the interactions. 
As the edges between nodes define interactions between pairs, our model utilizes the interaction proposal scores as the intensities of the edges in the graph. Interaction proposal scores are generated from the spatially refined visual features and they quantify if the human-object pair is interacting. 

To summarize, the proposed VSGNet for HOI detection refines the visual features using spatial relations of humans and objects. This approach amplifies the visual features of spatially relevant pairs while damping the others. Additionally, this model uses graph convolutional networks to model the interactions between humans and objects. The resulting model consists of multiple specialized branches. We evaluate our model on V-COCO\cite{gupta2015visual} and HICO-DET\cite{chao2018learning} datasets and demonstrate $4$ mAP ($~8\%$) and $3$ mAP ($~16\%$) improvement over the state of the art methods.  

\vspace{0.3cm}
\textbf{Technical Contributions:}
 \begin{itemize}
  \item We propose a new spatial attention branch that leverages the spatial configuration of human-object pairs and refines the visual features such that spatially relevant human-object pairs are amplified. 
  \item We use a graph convolutional branch which utilizes the structural connections between humans and objects. The interaction proposal score, generated from the spatially refined features, are used to define the edge intensities between human and object nodes.
  \item We implement a robust pipeline that contains Visual, Spatial and Graph based branches named VSGNet. This model achieves state-of-the-art results for HOI detection task on V-COCO and HICO-DET datasets. 
\end{itemize}

\section{Related Work}

{\bf Object Detection:} For detecting HOIs the first step is to detect humans and objects properly. With the recent object detection frameworks like RCNN~\cite{girshick2014rich}, Faster RCNN~\cite{ren2015faster}, YOLO~\cite{redmon2016you}, Feature Pyramid Network ~\cite{lin2017feature} and SSD~\cite{liu2016ssd}, models are able to detect multi scale objects robustly in images. Following this we utilize a pre-trained Faster-RCNN model in our network for detecting humans and objects. Additionally, we utilize the region proposal network idea from Faster-RCNN and extend it to interaction proposals which predict if an human-object pair is interacting.

{\bf Human Object Interaction:} Activity recognition is a research area in computer vision that has received interest for a long time. There are different datasets like UCF-101~\cite{soomro2012ucf101}, Thumos~\cite{idrees2017thumos} with a focus on detecting human actions in videos. But in these datasets, the goal is to detect one action in a short video which is not representative of real life scenarios. To extend human activity recognition in images Gupta et al.~\cite{gupta2015visual} introduced V-COCO dataset and Chao et al. ~\cite{chao2018learning} introduces HICO-DET dataset. These datasets are different from the previous datasets as they require models to explicitly detect humans, objects and their interactions. This extends the task to include detection of human activities while localizing the humans and the objects. 

For the HOI detection task, Gkioxari et al.~\cite{gkioxari2018detecting} proposed a human-centric approach arguing that human appearance provides strong cues in both detecting the action and localizing the object. This method does not consider interactions where the object is far away from the human. Qi et al.~\cite{qi2018learning} proposed a graph based network which depends on detecting an adjacency matrix between various nodes(here, nodes are humans and objects) but does not utilize any spatial relation cues between pairs. Kolesnikov et al.~\cite{kolesnikov2019detecting} incorporates the HOI detection process with the object detection by individually analyzing humans and objects without considering the relationship between the pairs.

Gao et al.~\cite{gao2018ican} proposed an attention network based on the previous work of~\cite{vaswani2017attention}. They derived an attention map from the human and object features over the whole convolutional feature map. Although they used a binary spatial map similar to~\cite{chao2018learning}, they use the spatial map to extract features and concatenate them with human visual features. 
As these are two completely different features defining separate things, concatenation does not enforce spatial configurations as much as an attention mechanism. 
To address this in our network we use the spatial features as attention maps which refines our visual features.

Li et al.~\cite{li2019transferable} integrated pose estimation with the iCAN~\cite{gao2018ican} and predicted the interaction probabilities between a human and object pair.  These methods however, do not explicitly leverage the interaction probabilities to detect the relational structure between the human and object pairs. Our VSGNet addresses this by utilizing a graph network for learning interactions and achieves better results without using poses which shows VSGNet can benefit from pose estimation as well. 
\section{Proposed Method}

\begin{figure*}[t]
\begin{center}
\includegraphics[width=1.0\linewidth]{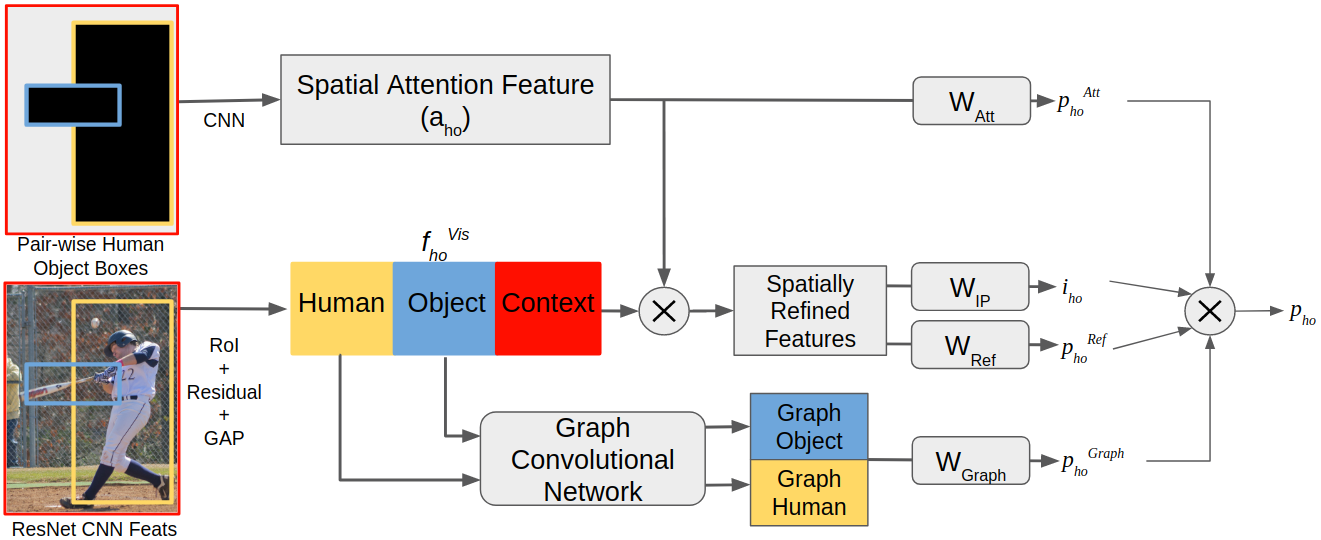}
\end{center}
   \caption{Model Architecture. Rounded rectangles are operations, sharp rectangles are extracted features and $\otimes$ is element-wise multiplication. The model consists of three main branches. Visual branch extracts human, object and context features. Spatial Attention branch refines the visual features by utilizing the spatial configuration of the human-object pair. Graph Convolutional branch extracts interaction features by considering humans/objects as nodes and their interactions as edges. Action class probabilities from each branch and the interaction proposal score are multiplied together to aggregate the final prediction. These operations are repeated for every human-object pair. \vspace{-0.5cm}
   }
\label{fig:model_architecture}
\end{figure*} 

This section introduces our proposed VSGNet for detecting human-object interactions(HOI). From each given image, the task is to detect bounding boxes for the humans, objects and correctly label the interactions between them. Each human-object pair can have multiple interaction labels and each scene can include multiple humans and objects in them. We simplify the task by running a pre-trained object detector which detects humans and objects in an image.

Detecting the interactions between human-object pairs is a challenging task. Simple methods such as extracting features from human and object locations individually and analyzing them are ineffective as these methods ignore the contextual information of the surroundings and the spatial relations of the human-object pair. Extensions such as using union boxes to model the spatial relations/context also fall short as they don't explicitly model the interactions. To address these issues, we propose a multi-branch network with specialized branches. The proposed VSGNet consists of the Visual Branch (Section \ref{sec:visual_branch}) which extracts visual features from human, object and surrounding context individually; the Spatial Attention Branch (Section \ref{sec:spatial_att}) which models spatial relations between the human-object pair; and the Graph Convolutional Branch (Section \ref{sec:graph_convs}) which considers the scene as a graph with humans and objects as nodes and models the structural interactions. The proposed model architecture with the branches is shown in Fig.\ref{fig:model_architecture}.

\subsection{Overview}

The inputs to our model is image features $\textbf{F}$ from a backbone CNN (e.g. ResNet-152~\cite{he2016deep}) and bounding boxes $x_h$ for human $h\in[1,H]$ and $x_o$ for object $o\in[1,O]$. $H$ and $O$ represents the total number of humans and objects in the scene respectively. Bounding boxes are obtained from a pre-trained object detector. 
We define the objective of this model as: 
\begin{itemize}
    \item Detect if human $h$ is interacting with object $o$ with an interaction proposal score $i_{h,o}$. 
    \item Predict the action class probability vector $\mathbf{p}_{h,o}$ of size $A$ where $A$ is the number of classes.
\end{itemize}
    

\subsection{Visual Branch} \label{sec:visual_branch}

This branch focuses on extracting visual features for the human-object pairs. Following the object detection methods, we use region of interest (RoI) pooling on the human/object regions to extract features. This operation is followed by a residual block (Res) \cite{he2016deep} and global average pooling(GAP) operations to extract the visual feature vectors for objects and humans. 

\begin{align}
    \mathbf{f}_h = GAP(Res_h(RoI(\mathbf{F}, x_h))) \\
    \mathbf{f}_o = GAP(Res_o(RoI(\mathbf{F}, x_o)))
\end{align}

\noindent where $Res_{\{\}}$ represents residual blocks, $\textbf{f}_h$ and $\textbf{f}_o$ are visual feature vectors of sizes $R$. This operation is repeated for each human $h$ and object $o$. 

Context plays an important role in detecting HOI. Surrounding objects, background and other humans can help detecting the interactions. We include the context in our network by extracting features from the entire input image followed by a residual block and global average pooling. 

\begin{equation}
    \mathbf{f}_C = GAP(Res_C(\mathbf{F}))
\end{equation}

\noindent where $\mathbf{f}_C$ is a feature vector of size $R$.

Finally, this branch combines all the visual feature vectors by concatenating them and projecting it by a fully connected layer.

\begin{equation}
    \mathbf{f}^{Vis}_{ho} = \mathbf{W}_{vis}(\mathbf{f}_h \oplus \mathbf{f}_o \oplus \mathbf{f}_C)
\end{equation}

\noindent where $\oplus$ is the concatenation operation,  $\mathbf{W}_{\{\}}$ is the projection matrix, $\mathbf{f}^{Vis}_{ho}$ is the combined visual feature vector of dimension $D$ which represents the human-object pair $ho$.

The feature $\mathbf{f}^{Vis}_{ho}$ can be used directly for classifying actions. We implement this as a base model for comparisons.

\subsection{Spatial Attention Branch} \label{sec:spatial_att}

\begin{figure}[t]
\begin{center}
\includegraphics[width=1.0\linewidth]{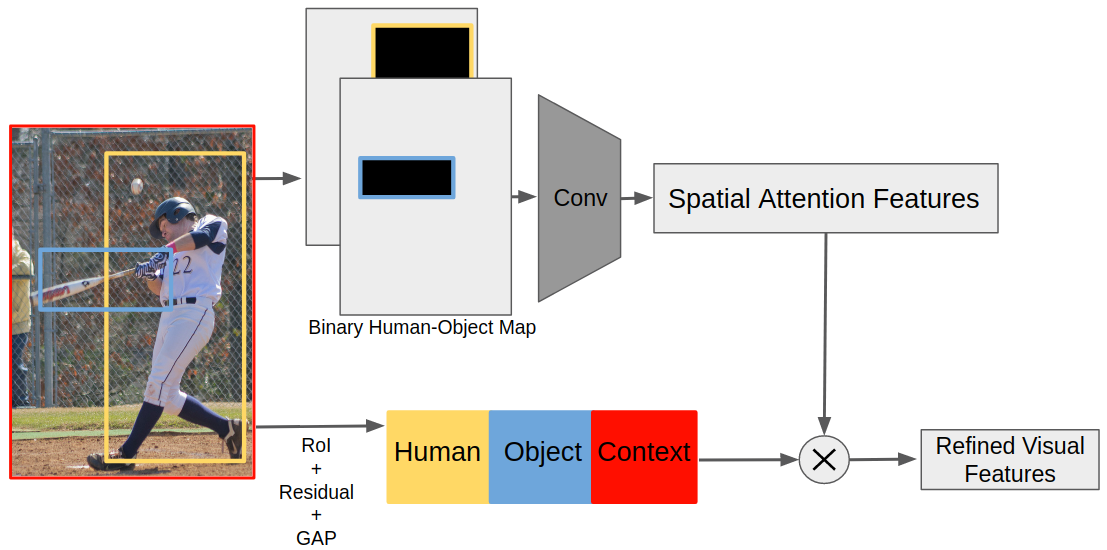}
\end{center}
   \caption{Spatial Attention Branch. Initially human, object and context visual features are extracted from the image using RoI pooling. Using binary maps of human and object locations, spatial attention features are extracted using convolutions. These attention features encode the spatial configuration of the human-object pair. Attention features are used to refine the visual features by amplifying the pairs with high spatial correlation. 
   }
\label{fig:spatial_attention}
\end{figure} 

This branch focuses on learning the spatial interaction patterns between humans and objects. The main task is to generate attention features which are used to refine the visual features by amplifying the pairs with high spatial correlation. This branch is visualized in Fig.\ref{fig:spatial_attention}.

Given the human bounding box $x_h$ and object bounding box $x_o$, we generate two binary maps. These binary maps have zeros everywhere except in locations defined by human and object box coordinates $x_h$ and $x_o$ for each map respectively. This generates a 2-channel binary spatial configuration map $\mathbf{B}_{ho}$. 

Similar to \cite{chao2018learning, gao2018ican}, we use 2 layers of convolutions to analyze the binary spatial configuration map. This is followed by a GAP operation and a fully connected layer. 

\begin{equation}
    \mathbf{a}_{ho} = \mathbf{W}_{Spat}(GAP(Conv(\mathbf{B}_{ho})))
\end{equation}

\noindent where $\mathbf{a}_{ho}$ is an attention feature vector of size D and represents the spatial configuration of the human-object pair $ho$. As the objects and humans are defined in different channels, using convolutions on the binary spatial configuration maps $\mathbf{B}_{ho}$ allows the model to learn the possible spatial relations between humans and objects.

Since $\mathbf{a}_{ho}$ encodes the spatial configuration, it can be used directly to classify the HOIs as in \cite{chao2018learning}. We keep this classification as an auxiliary prediction but mainly use $\mathbf{a}_{ho}$ as an attention mechanism for refining visual features. Auxiliary predictions can be defined as:

\begin{equation}
    \mathbf{p}^{Att}_{ho} = \sigma(\mathbf{W}_{Att}(\mathbf{a}_{ho}))
\end{equation}

\noindent where $\mathbf{p}^{Att}_{ho}$ is the action class probabilities of size $A$ and $\sigma$ is the sigmoid function.

The attention vector $\mathbf{a}_{ho}$ and the visual feature vector $\mathbf{f}^{Vis}_{ho}$ are set to be the same size $D$. This allows us to multiply these two vectors together in order to refine the visual features with spatial configuration. We use $\mathbf{a}_{ho}$ as an attention function and multiply $\mathbf{a}_{ho}$ and $\mathbf{f}^{Vis}_{ho}$ elementwise. 

\begin{equation}
    \mathbf{f}_{ho}^{Ref} = \mathbf{a}_{ho} \otimes \mathbf{f}^{Vis}_{ho}
\end{equation}

\noindent where $\otimes$ is element-wise multiplication and $f_{ho}^{Ref}$ is the spatially refined feature vector of size $D$. 

The refined feature vector is then used to predict the interaction proposal score of human-object pair $ho$ and to predict the action class probabilities. 

\begin{align}
    i_{ho} = \sigma(\mathbf{W}_{IP}(\mathbf{f}_{ho}^{Ref})) \\
    \mathbf{p}^{Ref}_{ho} = \sigma(\mathbf{W}_{Ref}(\mathbf{f}_{ho}^{Ref}))
\end{align}

\noindent where $i_{ho}$ is the interaction proposal probability of size 1 and $\mathbf{p}_{ho}^{Ref}$ is the action class probabilities of size $A$.

\subsection{Graph Convolutional Interaction Branch}\label{sec:graph_convs}

This branch uses a graph convolutional network to generate effective features for humans and objects. Graph convolutional networks extract features that model the structural relations between nodes. This is done by traversing and updating the nodes in the graph using their edges. In this setting, we propose to use humans and objects as nodes and their relations as edges.

\begin{figure}[t]
\begin{center}
\includegraphics[width=1.0\linewidth]{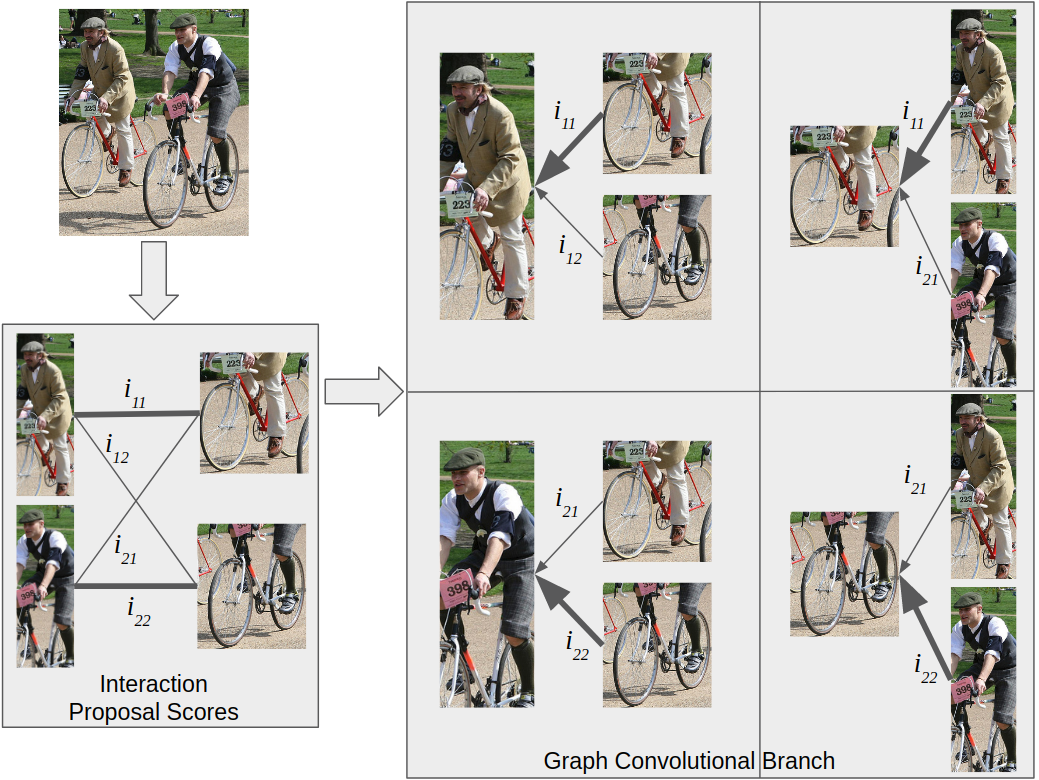}
\end{center}
   \caption{ Graph Convolutional Branch. This model learns the structural connections between humans and objects. For this task, we define the humans and objects as nodes and only connecting edges between human-object pairs. Instead of using visual similarity as the edge adjacency, we propose to use the interaction proposal scores. This allows the edges to utilize the interactions between human-object pairs and generates better features. 
   }
\label{fig:graph_convs}
\end{figure} 

Instead of having a fully connected graph, we connect each human with every object and each object with every human. 
However, without this simplification, proposed model can also be extended to fully connected settings. 

Given the visual features $\mathbf{f}_h$, $\mathbf{f}_o$ and connecting the edges between humans and objects, graph features $\mathbf{f}'_h$ and $\mathbf{f}'_o$ are defined as follows:

\begin{align}
    \mathbf{f}'_h = \mathbf{f}_h + \sum_{o=1}^{O} \alpha_{ho} \mathbf{W}_{oh}(\mathbf{f}_o) \\
    \mathbf{f}'_o = \mathbf{f}_o + \sum_{h=1}^{H} \alpha_{oh} \mathbf{W}_{ho}(\mathbf{f}_h)
\end{align}

\noindent where $\alpha_{ho}$ defines the adjacency between $h$ and $o$ and $\mathbf{W}_{oh}$, $\mathbf{W}_{ho}$ are mapping functions which project the object features to human feature space and vice versa. Previous works \cite{li2019relation, qi2018learning} defined the adjacency as visual similarity. In our task, however, adjacency defines interactions between nodes of visually unsimilar things which are human and object. Following this idea, we define adjacency values between $h$ and $o$ pair as:

\begin{equation}
    \alpha_{ho} = \alpha_{oh} = i_{ho}
\end{equation}

\noindent where $i_{ho}$ is the interaction proposal score which are generated from the spatially refined visual features and measure the interactions of the human-object pair. Pairing up the graph features, classification predictions are calculated as: 

\begin{equation}
    \mathbf{p}^{Graph}_{ho} = \sigma(\mathbf{W}_{graph}(f'_h \oplus f'_o))
\end{equation}

\noindent where $\oplus$ is concatenation operation and $\mathbf{p}^{Graph}_{ho}$ is the action class probabilities of size $A$. 

The graph convolutional branch is visualized in Figure \ref{fig:graph_convs}. This concludes all of the outputs of the proposed network. Finally we combine the action predictions and the interaction proposal scores by multiplying the probabilities similar to previous works \cite{gao2018ican, li2019transferable, gkioxari2018detecting}. 

\begin{equation}
    \mathbf{p}_{ho} = \mathbf{p}^{Att}_{ho} \times \mathbf{p}^{Ref}_{ho} \times \mathbf{p}^{Graph}_{ho} \times i_{ho}
\label{eq:infer}
\end{equation}

\noindent where $P_{ho}$ is the final prediction vector of size $A$.


\section{Experiments}

We first introduce the datasets and our evaluation metrics along with our implementation details and then perform extensive quantitative and qualitative analysis on our model and show the improvements over the existing methods.

\subsection{Datasets and Evaluation Metrics}

\textbf{Datasets:} To evaluate our model's performance, we use the V-COCO~\cite{gupta2015visual} and HICO-DET~\cite{chao2018learning} datasets. 

V-COCO is derived from COCO~\cite{lin2014microsoft} dataset. It has 10,346 images. 2533 images are for training, 2867 images are for validating and 4946 images are for testing. The training and validation set images are from COCO training set and the test images are from the COCO validation set. Each person in the images are annotated with a label indicating one of the 29 actions. If an object in the image is related to that action then the object is also annotated. Among these 29 actions, four of them has no object pair and one of them(point) has only 21 samples. Following the previous HOI detection works, we are not going to report our performance in these classes. We report our performance for the rest of the 24 classes.

HICO-DET is a large dataset for detecting HOIs with 38118 training and 9658 testing images. HICO-DET annotates the images for 600 human-object interactions. Following the previous works, in HICO-DET we report our performance in Full, Rare and Non-Rare Categories. These categories are based on the number of training samples ~\cite{chao2018learning}.   

\textbf{Metrics:}
Following~\cite{gupta2015visual} we evaluate our performance on two types of average precision(AP) metrics: Scenario 1 and Scenario 2. During AP calculation in both metrics, a prediction for a human-object pair is considered correct (1) if the human and object bounding boxes have an IoU greater than 0.5 with the ground-truth boxes and (2) the interaction class label of the prediction for the pair is correct. For the cases when there is no object(human only), in Scenario 1 a prediction is correct if the corresponding bounding box for the object is empty and in Scenario 2 the bounding box of the object is not considered. This makes Scenario 1 much harsher than Scenario 2~\cite{gupta2015visual}. In HICO-DET our evaluation metrics is similar to the Scenario 1 case of V-COCO.

\subsection{Implementation Details}
Resnet-152~\cite{he2016deep} network is used as the backbone feature extractor. We extract the input feature map before the last residual block of Resnet-152. This serves as the input to the rest of the network. We extract $10 \times 10$ feature maps for all the humans and objects from the input feature map by region of interest pooling~\cite{girshick2015fast}. Extracted RoIs and input feature map(context) pass through a residual block followed by a global average pooling similar to ~\cite{gao2018ican}. After these steps, we obtain three feature vectors of size $R=1024$ for human, object and context. These are fed to the rest of the network. For the spatial attention branch we have used $64 \times 64 \times 2$ binary inputs. Before the element wise multiplication with the attention vector in the spatial attention branch, we project all our input feature vectors to a $D=512$ dimensional space followed by a ReLU. In our final classification layer for all the branches, we have one linear layer.

For training the network, we utilize off-the-shelf Faster-RCNN~\cite{ren2015faster} to generate human and object bounding boxes. We have filtered the detected bounding boxes by setting 0.6 confidence threshold for human bounding boxes and 0.3 for object bounding boxes. The threshold values are chosen experimentally. Following ~\cite{gkioxari2018detecting} we did not fine tune the backbone CNN Resnet-152~\cite{he2016deep} and Faster-RCNN during our training process. Faster-RCNN was trained on the COCO \cite{lin2014microsoft} training set and did not see any image from V-COCO testing sets. Unlike previous works\cite{gao2018ican, li2019transferable}, we do not use ground truth boxes during training as object proposals. As our object detector is robust, we directly use the bounding boxes generated from the detector which generates sufficient amount of positive and negative boxes.


Initially, we have trained the model on the training set of V-COCO while validating with the validation set. Then we train the model in both training and validation set like~\cite{gkioxari2018detecting}. Our initial learning rate is set to 0.01 with a batch size of 8. As optimizer, Stochastic Gradient Descent(SGD) have been used with a weight decay of 0.0001 and a momentum of 0.9. To reduce the training time we have increased our learning rate to 0.01 for all the layers except for the spatial attention branch between epoch 9 to epoch 21. We trained the whole model for 50 epochs.

For HICO-DET we use the same hyper-parameters from V-COCO. We train the network individually for 20 epochs in HICO-DET training set without any initialization from the V-COCO model.  

During inference, we multiply all the prediction outputs from the different branches of our network as in \ref{eq:infer}. Additionally, we multiply the final prediction output with the detection confidences of the human and object from the object detector. To differentiate between high and low quality detection scores we have adopted Low grade Instance Suppressive Function (LIS)~\cite{li2019transferable}. We additionally remove the incompatible interaction-object pairs by using a post processing similar to iCAN ~\cite{gao2018ican} (e.g. if the object is not phone then the interaction can not be talk on the phone). 


While making inference most of the existing\cite{gao2018ican, li2019transferable, gkioxari2018detecting} models multiply all the outputs from different modules but these modules are optimized separately while training. Following ~\cite{gupta2019no} we have used a single cross entropy loss function for each action class to optimize the network. One thing to note is that as in Eq.~\ref{eq:infer}, interaction proposal score is also multiplied in these predictions and included in predictions for every class. This allows the proposal score to quantify if there are interactions between the human-object pair regardless of the class of that interaction. Our experiments show that combining all the predictions and using a single loss function improves the performance.


\subsection{Comparisons with the State of the Art}

We compare our model's performance with five recent state of the art methods~\cite{gkioxari2018detecting,kolesnikov2019detecting,qi2018learning,gao2018ican,li2019transferable} in both of the datasets. We report mean Average Precision (mAP) score in the settings provided by~\cite{gupta2015visual} and ~\cite{chao2018learning}.

Table \ref{tab:state_of_the_art_vcoco} shows that our method outperforms all the existing models and achieves an improvement of 4 mAP in scenario 1 for V-COCO dataset. We also reported our performance in scenario 2 which outperforms all the available existing methods who reported their results in that scenario.

\begin{table}[t]
\begin{center}
\begin{tabular}{|l|c|c|}
\hline
V-COCO & mAP(Sc 1) & mAP(Sc 2)\\
\hline\hline

InteractNet~\cite{gkioxari2018detecting} & 40.0 & 47.98\\\hline
Kolesnikov et al.~\cite{kolesnikov2019detecting} & 41.0 & -\\\hline
GPNN~\cite{qi2018learning} & 44.0 & - \\\hline
iCAN~\cite{gao2018ican} & 45.3 & 52.4\\\hline
Li et al.~\cite{li2019transferable} & 47.8& -\\\hline\hline
VSGNet & {\bf 51.76} &{\bf 57.03}\\\hline

\end{tabular}
\end{center}
\caption{Comparison of results in V-COCO ~\cite{gupta2015visual} test set on Scenario 1 and Scenario 2. Our method outperforms the closest method by 8\%. For actor only classes (no object), scenario 1 requires the model to detect it specifically as no object, whereas scenario 2 ignores if there is an object assigned to that prediction. Some of these methods did not provide results for scenario 2.   
\vspace{-0.3cm}
}
\label{tab:state_of_the_art_vcoco}
\end{table}

Table \ref{tab:state_of_the_art_hico} shows the results compared to other methods in HICO-DET and our model achieves the best results among the previous works.

\begin{table}[t]
\begin{center}
\begin{tabular}{|l|c|c|c|}
\hline
HICO-DET (mAP) & Full & Rare & Non-Rare\\
\hline\hline
HO-RCNN~\cite{chao2018learning} &7.81 & 5.37 & 8.54\\\hline
InteractNet~\cite{gkioxari2018detecting} & 9.94  & 7.16  & 10.77\\\hline
GPNN~\cite{qi2018learning} & 13.11 & 9.34  & 9.34 \\\hline
iCAN~\cite{gao2018ican} & 14.84 & 10.45 & 16.15\\\hline
Li et al.~\cite{li2019transferable} & 17.03 & 13.42  & 18.11\\\hline\hline
{\bf VSGNet} & {\bf 19.80} &{\bf 16.05} &  {\bf 20.91}\\\hline

\end{tabular}
\end{center}
\caption{Comparison of results in HICO-DET ~\cite{gupta2015visual} test set. VSGNet outperforms the closest method by 16\%.
}
\label{tab:state_of_the_art_hico}
\end{table}

\begin{table}[t]
\begin{scriptsize}
\begin{center}
\begin{tabular}{|c|c|c|c|}
\hline
HOI Class        & InteractNet~\cite{gkioxari2018detecting} & iCAN~\cite{gao2018ican} & VSGNet                            \\ \hline
hold-obj         & 26.38       & 29.06                                         & \textbf{48.27} \\ \hline
sit-instr        & 19.88       & 26.04                                         & \textbf{29.9}                            \\ \hline
ride-instr       & 55.23       & 61.9                                          & \textbf{70.84} \\ \hline
look-obj         & 20.2        & 26.49                                         & \textbf{42.78} \\ \hline
hit-instr        & 62.32       & 74.11                                         & \textbf{76.08} \\ \hline
hit-obj          & 43.32       & 46.13                                         & \textbf{48.6}                           \\ \hline
eat-obj          & 32.37       & 37.73                                         & \textbf{38.3} \\ \hline
eat-instr        & 1.97        & \textbf{8.26}                                          & 6.3                            \\ \hline
jump-instr       & 45.14       & 51.45                                         & \textbf{52.66}                           \\ \hline
lay-instr        & 20.99       & \textbf{22.4}                                          & 21.66                           \\ \hline
talk\_on\_phone  & 31.77       & 52.81                                         & \textbf{62.23} \\ \hline
carry-obj        & 33.11       & 32.02                                         & \textbf{39.09}                           \\ \hline
throw-obj        & 40.44       & 40.62                                         & \textbf{45.12}  \\ \hline
catch-obj        & 42.52       & \textbf{47.61}                                         & 44.84                           \\ \hline
cut-instr        & 22.97       & 37.18                                         & \textbf{46.78} \\ \hline
cut-obj          & 36.4        & 34.76                                         & \textbf{36.58} \\ \hline
work\_on\_comp   & 57.26       & 56.29                                         & \textbf{64.6}                           \\ \hline
ski-instr        & 36.47       & 41.69                                         & \textbf{50.59}                           \\ \hline
surf-instr       & 65.59       & 77.15                                         & \textbf{82.22} \\ \hline
skateboard-instr & 75.51       & 79.35                                         & \textbf{87.8} \\ \hline
drink-instr      & 33.81       & 32.19                                         & \textbf{54.41} \\ \hline
kick-obj         & 69.44       & 66.89                                         & \textbf{69.85} \\ \hline
read-obj         & 23.85       & 30.74                                         & \textbf{42.83} \\ \hline
snowboard-instr  & 63.85       & 74.35                                         & \textbf{79.9} \\ \hline\hline
Average          & 40.0           & 45.3                                          & \textbf{51.76} \\ \hline
\end{tabular}
\end{center}
\caption{Per class AP comparisons to the existing methods in V-COCO Scenario 1. Our method demonstrates superior performance in majority of the classes. We only compared to the methods which have reported the per class AP values. Obj refers object cases where instr refers to instrument~\cite{gupta2015visual}.
\vspace{-0.3cm}
}
\label{tab:per class Performarmance}
\end{scriptsize}
\end{table}

The closest work to our results is Li et al.~\cite{li2019transferable} which builds on top of iCAN \cite{gao2018ican} by adding an interaction proposal network and utilizing person poses. Addition of interaction proposal and person poses improve $\sim2$ mAP in V-COCO and $\sim3$ mAP in HICO-DET on top of iCAN with a computational cost of calculating the poses for each human. Our model achieves better results without the pose extraction and can possibly improve another 5\% if the pose features are added to our visual feature branch. 

In Table \ref{tab:per class Performarmance} we report per-class performances compare with the existing methods which reported per-class APs for V-COCO. Our proposed VSGNet achieves better performance in majority of the classes compared to the other methods.
Additionally, per-class performances show that some of the action classes perform badly due to the failure of object detectors (e.g. eat instruments which usually have small objects and commonly become occluded in the images). As our main task is to detect HOIs, we did not fine-tune the existing object detectors according to our needs which can also possibly handle these cases.

\subsection{Ablation Studies}

\noindent\textbf{Analysis of Individual Branches:}
Our overall architecture consists of three main branches.
To evaluate how these branches are affecting our overall performance, we evaluate these branches individually in the V-COCO ~\cite{gupta2015visual} test set. Our evaluation method and metrics are same as Table \ref{tab:state_of_the_art_vcoco}. We consider the base model as the Visual branch without the spatial attention or the graph convolutions. In this setting, interaction proposal score $I_{ho}$ and the class probabilities $P_{ho}$ are predicted from the visual features $f^{Vis}_{ho}$ directly. 

We have added the graph network and the spatial network with our base model individually to evaluate each of the branch's performance separately. The results are shown in Table \ref{tab:ablation}. With addition of the individual branches, model performance has improves gradually. Visual+Spatial branch achieves state of the art results by itself without the Graph branch. Addition of the graph branch adds additional 1.5 mAP and a total of 4mAP over the state of the art.

An important detail is that the graph branch directly depends on the quality of the interaction proposal score $i_{ho}$ as it is used to determine the edge interactions. Without the spatial attention, visual features generate inferior $i_{ho}$ which affects the graph branch. This is the reason that addition of Graph to Visual branch only adds 0.9 mAP whereas addition of Graph to Visual+Spat makes a larger improvement and adds 1.5 mAP.

Spatial attention branch improves the result by 3 mAP when added to the visual branch. This demonstrates the importance of the spatial reasoning and refining the visual features. Graph and Spatial attention combined improves the performance by about 4.5 mAP over the base model. 

\begin{table}[t]
\begin{center}
\begin{tabular}{|l|c|c|}
\hline
Branches & mAP(Sc 1) & mAP(Sc 2)\\
\hline\hline

Visual (Base) & 47.3 & 52.15\\\hline
Visual+Graph & 48.19 & 53.12\\\hline
Visual+Spatial & 50.33 & 55.32 \\\hline
Visual+Spatial+Graph(VSG) & 51.76 & 57.03\\\hline

\end{tabular}
\end{center}
\caption{ Analysis of the branches. Our base model consists of only the Visual branch. We add the graph branch and the spatial attention branch to this base model separately to analyze their performances. Individually, both branches improve the performance upon the base model. Visual+Spatial model beats the state of the art results and all three branches combined adds another 1.5 mAP. }
\label{tab:ablation}
\end{table}


\noindent\textbf{Analysis of Backbone CNNs:}
In addition to all Resnet models \cite{he2016deep}, we implement our model with various common CNNs used in image analysis. Table \ref{tab:resnet} shows the results of VSGNet implemented with these various backbone CNNs in V-COCO with Resnet152 performing the best.


\begin{table}[t]
\begin{center}
\begin{tabular}{|l|c|c|}
\hline
Branch & mAP (Scenario 1)  \\
\hline\hline

VGG-19\cite{simonyan2014very} & 48.37 \\\hline
InceptionV3\cite{szegedy2016rethinking} & 49.39 \\\hline
SqueezeNet\cite{iandola2016squeezenet} & 43.4 \\\hline
\hline

Resnet34\cite{he2016deep} & 50.88 \\\hline
Resnet50\cite{he2016deep} & 51.01 \\\hline
Resnet101\cite{he2016deep} & 50.01  \\\hline
Resnet152\cite{he2016deep} & \textbf{51.76}\\\hline

\end{tabular}
\end{center}
\caption{ Effects of the backbone CNN on V-COCO dataset. VSGNet is implemented using various common backbone CNNs. Resnet-152 model with VSGNet achieves the best performance. \vspace{-0.5cm}}
\label{tab:resnet}
\end{table}

\begin{figure*}[t]
\begin{center}
\includegraphics[width=0.8\linewidth]{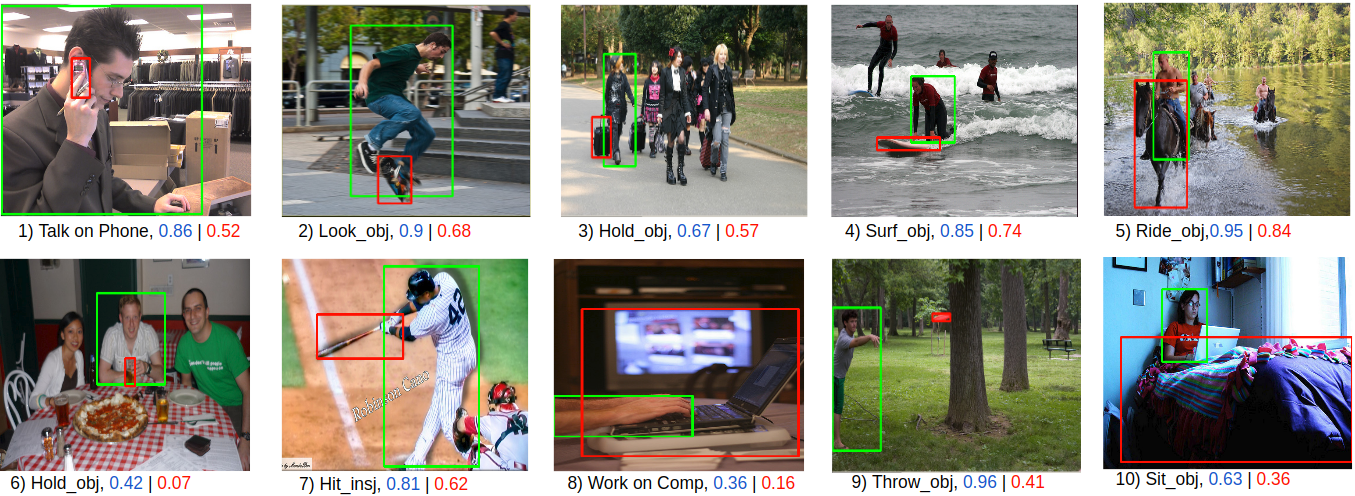}
\end{center}
   \caption{ Qualitative results. Red values show the confidences for the base model (Visual only) and blue values are the results for the VSGNet. The prediction results and the correct action labels are shown for the human-object pair visualized with the bounding boxes. 
   \vspace{-0.3cm}
   }
\label{fig:qualitative_res}
\end{figure*}

\noindent\textbf{Qualitative Results:}
Figure \ref{fig:qualitative_res} shows qualitative results and compares the VSGNet with the base model (Visual only). The interaction prediction probabilities for the correct action is visualized.  The images show the variance in object sizes, human sizes and different interaction classes. VSGNet performs better than the base model. Even in the cases when the object is not entirely visible (image 9) or the interaction is very subtle (image 2) VSGNet performs well and improves upon the base model.  

\begin{figure}[t]
\begin{center}
\includegraphics[width=0.75\linewidth]{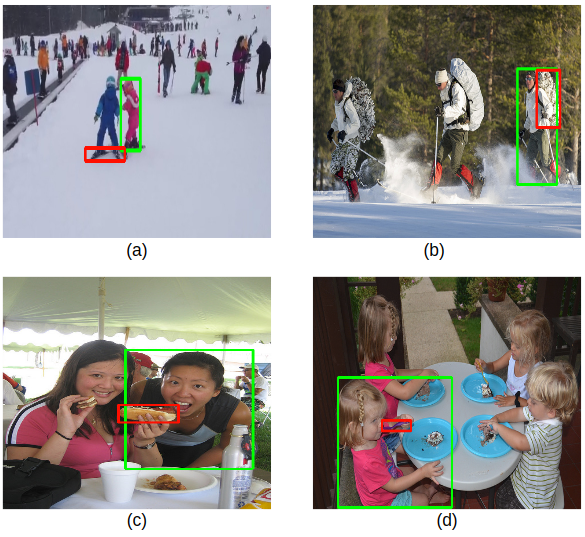}
\end{center}
   \caption{ Few of the cases where our VSGNet's prediction is wrong due to the confusing visual and spatial cue from the images. (a) Human-object pair is detected to be interacting but they are not, (b) Label mismatch (hold vs carry), (c) confusing scene  and (d) object detector fails to detect the fork. 
   \vspace{-0.3cm}
   }
\label{fig:failure_cases}
\end{figure}

\noindent\textbf{Failure Cases:}
When the visual or spatial cues are confusing, the model can fail to predict the correctly. In Figure \ref{fig:failure_cases} a few failure cases are shown. Our method can fail if the spatial configuration is confusing (a), confusing ground truth labels (hold and carry in (b)), multiple humans interacting with the same object with similar spatial configuration (c), the object detector fails to detect the objects of interest (d). 


\section{Discussions}

\subsection{Differences with similar works}

We compare VSGNet with methods using spatial relations \cite{chao2018learning, gao2018ican, li2019transferable}, attention\cite{gao2018ican} and graph convolutions\cite{qi2018learning, li2019relation}. 

There have been previous works which use spatial relation maps such \cite{chao2018learning, gao2018ican, li2019transferable}. These methods have either used the spatial relation maps directly for classification \cite{chao2018learning} or concatenated the spatial relation features to their visual features\cite{gao2018ican, li2019transferable}. Directly using them for classification ignores the visual features which in turn only learns relationship between the interaction label and spatial configuration. Concatenation of visual and spatial relations is also inferior to our method. 
As these are two completely different features defining separate things, concatenation does not enforce spatial configurations as much as an attention mechanism. 
In contrast, we use the spatial relations to extract attention features which are then used to alter the visual features. This is more effective as it models the relations between the visual feature channels and spatial configuration due to the element-wise multiplication. 

Attention models also have been used on HOI task. iCAN\cite{gao2018ican} model uses an attention model inspired from \cite{vaswani2017attention} and models the attention of the human or object region with the whole input scene individually. However, this approach does not consider the relation between the pairs and they only include the spatial configuration at the end. Our approach uses the spatial configuration directly to alter the visual features of the pairs which amplifies connected ones and dampens irrelevant ones at feature level. 

Graph convolutions \cite{qi2018learning, li2019relation} have been effective in various tasks. These tasks learn or use visual similarity as adjacency values between nodes and extract graph features. However, for our task, interaction proposal scores already defines the adjacencies between human-object node pairs and are used as edge intensities. This approach effectively extracts graph features by traversing relevant object nodes for the humans and relevant human nodes for objects.

\subsection{Summary}

We presented a novel human-object interaction detection model VSGNet which utilizes Visual, Spatial and Graph branches. VSGNet generates spatial attention features which alter the visual features and uses graph convolutions to model the interactions between pairs. The altered visual features generate interaction proposal scores which are used as edge intensities between human-object node pairs. We demonstrated with thorough experimentation that VSGNet improves the performance and outperforms the state-of-the-art.

\footnotesize{
\noindent\textbf{Acknowledgements:}
This work is partially supported by NSF SI2-SSI award \#1664172.  
}

{\small
\bibliographystyle{ieee_fullname}
\bibliography{main}
}

\end{document}